# Visual Semantic Multimedia Event Model for Complex Event Detection in Video Streams

Piyush Yadav and Edward Curry

Lero-Irish Software Research Centre,
National University of Ireland Galway, Ireland
{piyush.yadav,edward.curry}@lero.ie

**Abstract.** Multimedia data is highly expressive and has traditionally been very difficult for a machine to interpret. Middleware systems such as complex event processing (CEP) mine patterns from data streams and send notifications to users in a timely fashion. Presently, CEP systems have inherent limitations to process multimedia streams due to its data complexity and the lack of an underlying structured data model. In this work, we present a visual event specification method to enable complex multimedia event processing by creating a semantic knowledge representation derived from low-level media streams. The method enables the detection of high-level semantic concepts from the media streams using an ensemble of pattern detection capabilities. The semantic model is aligned with a multimedia CEP engine deep learning models to give flexibility to end-users to build rules using spatiotemporal event calculus. This enhances CEP capability to detect patterns from media streams and bridge the semantic gap between highly expressive knowledge-centric user queries to the low-level features of the multimedia data. We have built a small traffic event ontology prototype to validate the approach and performance. The paper contribution is threefold- i) we present a knowledge graph representation for multimedia streams, ii) a hierarchal event network to detect visual patterns from media streams and iii) define complex pattern rules for complex multimedia event reasoning using event calculus.

**Keywords:** Multimedia Event Model, Complex Event Processing, Computer Vision, Multimedia Stream Representation, Deep Neural Network, Traffic Surveillance, Event Calculus

## 1 Introduction

Event processing is an effective and scalable way to disseminate information to the intended user as they are open, distributed, and decoupled. These systems provide fast reasoning over data streams with high throughput and low latency to address velocity, variety and volume dimensions of big data in real-time. During the last decade, Complex Event Processing (CEP) systems have been increasingly adopted for real-time analysis in different domains including traffic, maritime surveillance, environmental and financial applications. CEP systems combine individual atomic events from streams to high-level semantics (complex event) and notify to interested users as per their query [1].

Due to recent advances in digital and sensor technology, there is a significant shift in the data landscape. We are now transitioning to an era of the Internet of Multimedia Things (IoMT) [2], where visual sensors prevail everywhere and are generating unprecedented volume of multimedia data. A recent report from EMC estimated a yearly growth of 42.5% of internet traffic with nearly 70% of it as multimedia data. Multimedia data present their own challenges, like an image can be interpreted in multiple ways depending on human perception (see Fig. 1), have a high degree of information (multiple objects and relationships) and are represented as complex low-level features. Thus, there is a need to overcome these limitations to build media processing capabilities in present CEP systems. In this context, we propose a multimedia stream enabled CEP with the following main contributions:

1. We present a semantic representation of multimedia streams in terms of knowledge graph using our proposed *Visual Concept Ontology for Multimedia events ($V_{COM}$)*.
2. We introduce *Multimedia Event Relation Network (MERN)*, a formal multimedia event model for visual pattern recognition using a top-down layered approach. The model captures the semantic hierarchies between various multimedia objects and their relationship in a domain and thus enable users to write queryable, actionable and explicit complex CEP queries for multimedia data.
3. We define event pattern rules using event calculus which can handle spatiotemporal relationship over multimedia streams thus enabling visual semantic queries in CEP.

The rest of the paper is organised as follows. Section 2 explains background and motivation, and Section 3 throws light on related work. Section 4 introduces the proposed approach for multimedia event modelling. Section 5 explains event detection rules and query formulation while system architecture and experimental results and limitations are explained in section 6. The paper discusses future work and conclusion in section 7.

## 2 Background and Motivation

### 2.1 Present CEP Limitations

Presently, CEP engines have various limitations to process multimedia data due to following reasons:

- *Unstructured Event Representation:* As shown in Fig. 1, an image is represented as low-level features (pixel values) to machine (CEP matcher) while human interprets them as high-level semantic labels (‘car’, ‘red car’) which creates a semantic gap between the user and the data. Present CEP engines (Esper, WSO2 CEP, Siddhi, Cayuga) [1] are designed using a fixed data model with a structured payload like key-value pairs, XML data and RDF triples. Reasoning over a structured data model with well-defined semantics and representation is a well-understood problem. Since multimedia streams have no structured representation, it is difficult to 1) express a pattern and 2) process the data to detect the pattern.

- *Fixed Event Model:* As shown in Fig. 1, in present CEP systems the event model is well-defined, and their process models have fixed rules and operators to detect complex events. Implementing a multimedia event model is challenging as the system would need to learn a very complex model which includes varied spatial, temporal, objects and structural information.
- *Event Pattern Rules:* It is difficult to create event pattern rules within multimedia streams as they can have a very dynamic nature where objects are in motion and can generate complex and changing relationships.

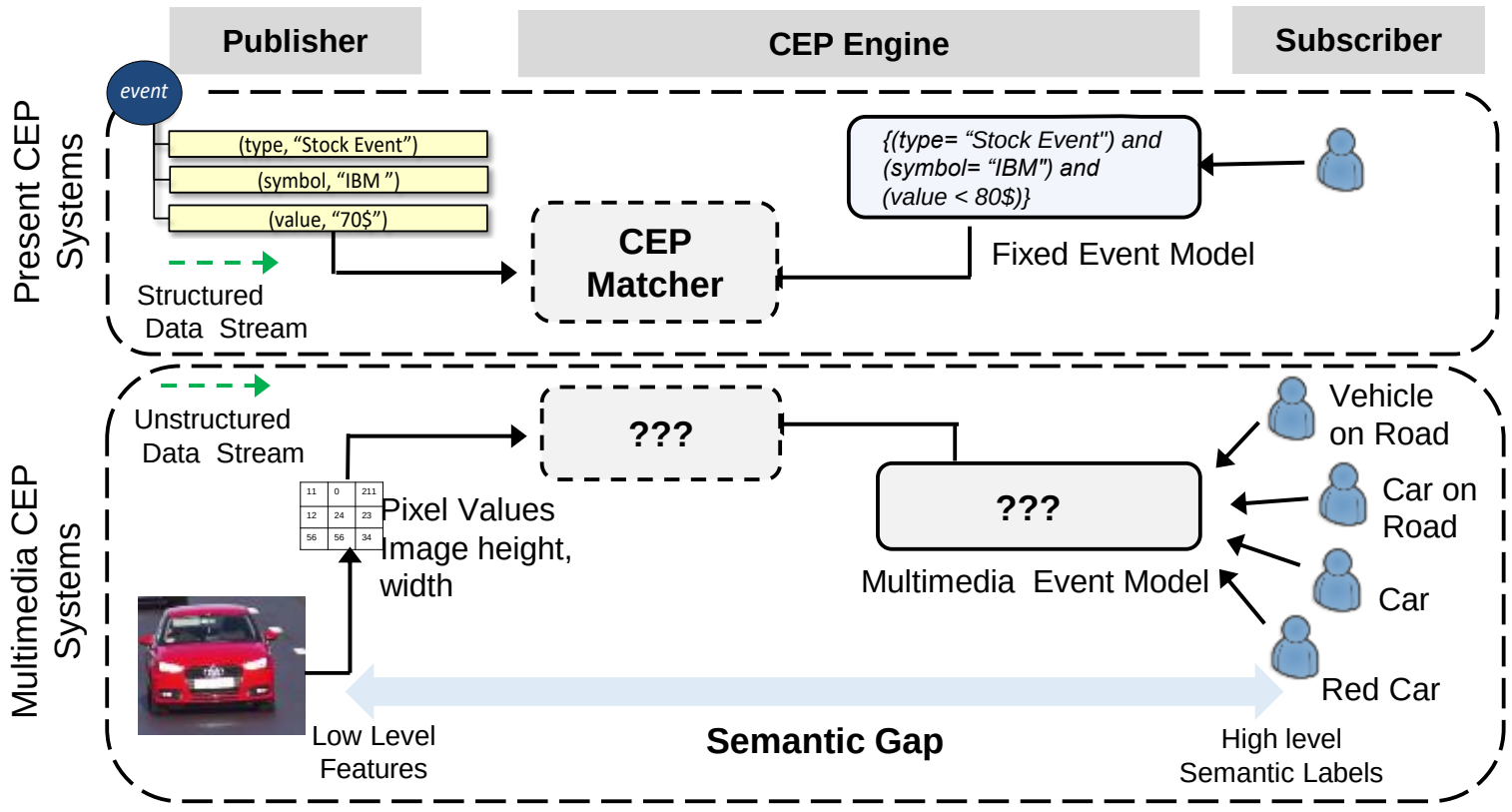

**Fig. 1.** Challenges in Multimedia Complex Event Processing

## 2.2 Motivational Scenario

Let's understand the problem from the perspective of the smart city traffic surveillance system. Suppose traffic police has declared a route of the city as 'no overtaking area'. They want to subscribe for an automated 'overtake' violation from traffic camera feeds using multimedia CEP engine where a vehicle should not pass another vehicle. Here, 'overtake' is a complex event which constitutes multiple atomic events like object(vehicle) detection, its position with respect to another object at different time steps etc. To process this query over video streams raises the following challenges:

1. How to represent low-level incoming media stream of media objects at different time instances?
2. How to write complex spatiotemporal pattern rules like 'overtake' in CEP engine?
3. How can a user define high-level knowledge-centric queries instead of focusing on system low-level logic?
4. How can the CEP engine match low-level media data with high-level knowledge-centric queries?

To overcome the above challenges, we need to bridge the semantic gap between low-level media features and high-level user queries to make a CEP engine to match a requested pattern (like ‘Overtake’). Semantic models like ontologies can establish connections between these entities by integrating semantic features and different levels of abstraction. This type of expressive formal representation has enabled reasoning over streams by mapping enriched background knowledge. Thus, we need to introduce a novel stream representation and ontology method within multimedia CEP which can detect complex patterns in multimedia data streams.

## 3 Related Work

***Deep Neural Network and their Limitation:*** Recently, Deep Neural Networks (DNN) have become a state-of-the-art method to solve problems related to multimedia data with good accuracy and performance. DNN based object detection models like YOLO [3] provide bounding boxes across the objects in the images which are highly accurate. Similarly, DNN based relation detection models [4] describe the relationship between the objects, but these models’ are better suited for static data like images where relationships are annotated among objects manually, and the model is trained to detect relationship among objects. Although DNN’s are quite promising they have certain limitations with respect to CEP systems which are listed below:

- *Query Dynamicity:* In DNN a system needs to be trained in an end to end fashion to detect a pattern. In CEP systems there may be different continuous user queries at a different time instance. It’s very challenging to train every pattern where requirement changes due to subscriber’s query dynamicity.
- *High Training and Computation Cost:* Training each pattern is costly in terms of resources and computation, and this is difficult in the highly dynamic and unpredicted environment.
- *Training Data Limitation:* DNN models’ performance is highly restricted with the amount of training data. In the visual world, there are infinite objects and relationships, and thus it’s nearly impossible to create a dataset for each pattern.

***CEP Languages for Multimedia Reasoning:*** Most of the CEP languages reason over structured data and are not suitable for processing multimedia streams. CEP systems use relational stream processing languages and work with the assumption that the incoming data has a fixed format. Taylor et al. [5] used ontology-driven CEP and were focused on multimedia sensor networks. They added only structured sensor information rather than the content of the image. Table 1 shows the comparison of various CEP and Multimedia retrieval languages. It can be seen that no CEP language has the sufficient capability to process multimedia streams.

**Table 1.** Scope of CEP Languages for Multimedia Data Streams

| CEP Language | | | | |
|---|---|---|---|---|
| Languages | Multimedia Content Extraction | Query Multimedia Content | Multimedia Streams | Multimedia Function Extension |
| Snoop | X | x | X | X |
| ESPER | X | x | X | X |
| SPARQL | X | x | X | X |
| ETALIS | X | x | X | X |
| EP-SPARQL | X | x | X | X |
| SPARQL-ST | X | x | X | X |
| Multimedia Retrieval Language | | | | |
| CVQL | X | Yes | Video Database | X |
| SVQL | X | Yes | Video Database | X |
| SPARQL-MM | X | Yes | Annotated Video | X |

***Multimedia Data Modelling:*** Initially, Westermann et al. proposed a theoretical high-level event model for multimedia applications[6]. In IMGpedia [7] authors added low-level features of the image to create a linked dataset of images, but they have captured no semantic relationship among them. In OVIS [8] authors have developed video surveillance ontology for large volumes of video in databases with no support for streaming. Xu et al. [9] presented a Video Structural Description(VSD) technology which is a generalised model for discovering semantic concepts and their relation in the video, with a focus on object search instead of pattern detection.

***Multimedia Event Detection:*** Initial work by Medioni et al. [10] was focused on detecting and tracking of moving objects using low-level image features. In 'REMIND' [11] Dubba et al. used Inductive Logic Programming to create relation event models for video. Our work overlaps with them regarding designing patterns using spatiotemporal calculus but differs regarding the use of CEP and multimedia event model. In [12] author created a framework 'Eventshop' focussing on situation recognition in multimedia data which aggregates data from various streaming heterogenous sources.

***Image Understanding:*** Chen et al. proposed an Object Relation Network (ORN) which is a guide ontology to recognise the scene in an image [13]. ORN acts as background knowledge to transfer rich semantics to identify relationships among objects based on an energy function. In KGA-CGM [14] Mogadala et al. have presented a knowledge-guided image captioning approach using knowledge graphs. Guangnan et al. proposed EventNet [15], a video event ontology with large-scale concept library based on WikiHow articles to match queries with the semantically relevant concept.

Inspired by the computer vision world, we aim to build semantic representations of multimedia streams using multimedia event model which will make it simpler for the CEP engine to reason over incoming media streams. This will enable the user to query interesting event pattern based on semantically enriched multimedia event model and the same can be reused among similar interested users. Thus, we introduce a novel stream representation and event model in our multimedia CEP engine which can detect complex patterns in multimedia data streams.

## 4 Approach for Multimedia Event Modelling

In order to detect patterns from multimedia streams our approach is as follows- i) First, we propose a Visual Concept Ontology ($V_{COM}$) to formalize the multimedia data streams, ii) Second, we introduce a multimedia event model MERN, a hierarchal semantically enriched layered network to enable pattern detection with a prototype example, iii) Third, we present a reference architecture for our multimedia CEP, 4) Finally, we implement different event pattern rules using spatiotemporal event calculus and queries to detect complex pattern from the streams.

### 4.1 Visual Concept Ontology for Multimedia Stream Representation ($V_{COM}$)

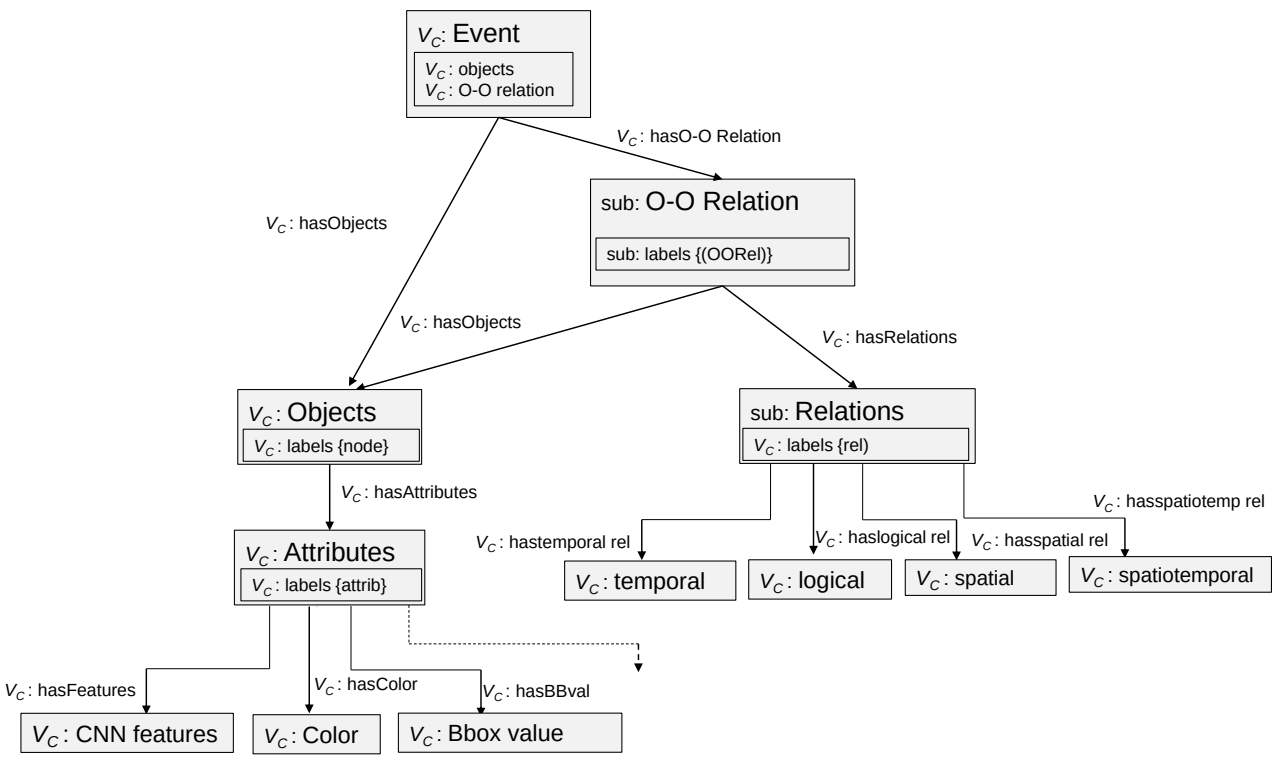

**Fig. 2.** Visual Concept Ontology (VCOM)

Extracting semantic information from multimedia data is a challenging task. Object detection techniques are not enough to define the complex relationships and interactions between the objects and thus limits the semantic expressiveness of data. Multimedia events have multiple visual artefacts relating from objects, attributes and relationships. These relationships can be– i) Temporal: Object to Object relation across time, ii) Spatial: Object to Object relation within images and iii) Spatiotemporal: Object to Object relation within and across images. Thus, there is a need to develop specifications which can incorporate visual information in our multimedia event model. As shown in Fig. 2 we propose a lightweight multimedia event representation in terms of ontological entities referred as Visual Concept Ontology for Multimedia Events ($V_{COM}$). $V_{COM}$ defines a multimedia event as the combination of objects (O) and object-object(O-O) relation as explained below:

- *Objects:* An object is constituted as a basic building block for any image or video. Thus, 'Objects' is one of the main class in $V_{COM}$. Objects can have multiple attributes like shape, colour, type, features (low-level) which are represented by the 'Attributes' subclass.
- *Relations:* The 'Relation' class defines the interaction among objects. These interactions can be at a spatial and temporal level which can be built using spatiotemporal event calculus.
- *O-O Relation:* The 'O-O' relation class consists of the 'Objects' and 'Relations' class. This class abstracts how two objects interact with each other. For example, in Overtake(Car, Bike), 'Overtake' is an O-O relation which establishes a spatio-temporal relation between two objects (O): 'Car' and 'Bike'.
- *Event:* An event is a superclass which consists of Object and O-O Relation class. Therefore, a user can query by providing an event class.

Thus, we define a multimedia event as- $MM\ Event = \{Objects, O - O\ Relations\}$. For example, a user want to query for a 'Black SUV Car' from a data stream so it can be simply written as: MM Event = $\{Car_{\text{Color: Black, Type: SUV}}, \emptyset\}$ where 'Car' is an object class with Attributes Type: SUV and Color: Black as its subclasses. Here Ø represents that user has not subscribed for any O-O relation class. Similarly, if user query is $MM\ Event = \{\emptyset, Overtake(Car, Bike)\}$, then here 'Overtake' is an O-O relation while 'Car' and 'Bike' are Objects on which they want to apply this relation. We can now represent multimedia stream as a time stamped multimedia event knowledge graph following the $V_{COM}$ structure which is defined below:

**Definition1:** A Multimedia Event Knowledge Graph ($ME_{KG}$) is a labelled multigraph with 5 tuples represented as - $\mathbf{ME_{KG} = \{O, O\text{-}O, A_V, R_{O\text{-}O}, \lambda\}}$ where $\boldsymbol{O}$ = set of object nodes, $\boldsymbol{O-O}$ = set of relation edges such that $\boldsymbol{O-O} \subseteq \mathbf{O}\ \chi\ \boldsymbol{O}$, $\boldsymbol{Av}$= set of attributes to each object class , $\boldsymbol{R_{O-O}}$= set of relations classes where $\boldsymbol{R_{O-O}} \cap \boldsymbol{O_v} = \boldsymbol{\phi}$ and $\lambda$ is a class labelling function over $\boldsymbol{O}$ and $\boldsymbol{O-O}$ classes. Presently, in $ME_{KG}$ the O-O relation is the spatial and temporal relation between the objects.

**Definition2:** A Multimedia Event Knowledge Graph Stream (S) is a sequence ordered representation of $ME_{KG}$ such that $\boldsymbol{S} = \{(\boldsymbol{ME_{KG}^1, t_1}), (\boldsymbol{ME_{KG}^2, t_2}) \dots \dots \dots (\boldsymbol{ME_{KG}^n, t_n})\}$ where $t\epsilon\ timestamp$ such that $t_i < t_{i+1}$.

## 4.2 Multimedia Event Relation Network (MERN)

Leveraging $ME_{KG,}$ we have created a knowledge graph representation for multimedia data. Now the questions come how to define a semantic event pattern like 'Overtake' in multimedia CEP which is across space and time. Thus, we propose a hierarchal Multimedia Event Relation Network (MERN) model which recognises event patterns between media objects based on their spatial and temporal layouts. MERN helps in creating a predefined ontology which includes prebuild semantic relationships between objects with the help of an event calculus. This helps users to write queryable patterns over incoming media streams. The MERN aligns with the systems deep learning model detection capabilities and is used to create an enriched semantic knowledge of a domain. Fig. 3 shows an example of MERN model which is hierarchically divided into five layers to encode traffic related multimedia events:

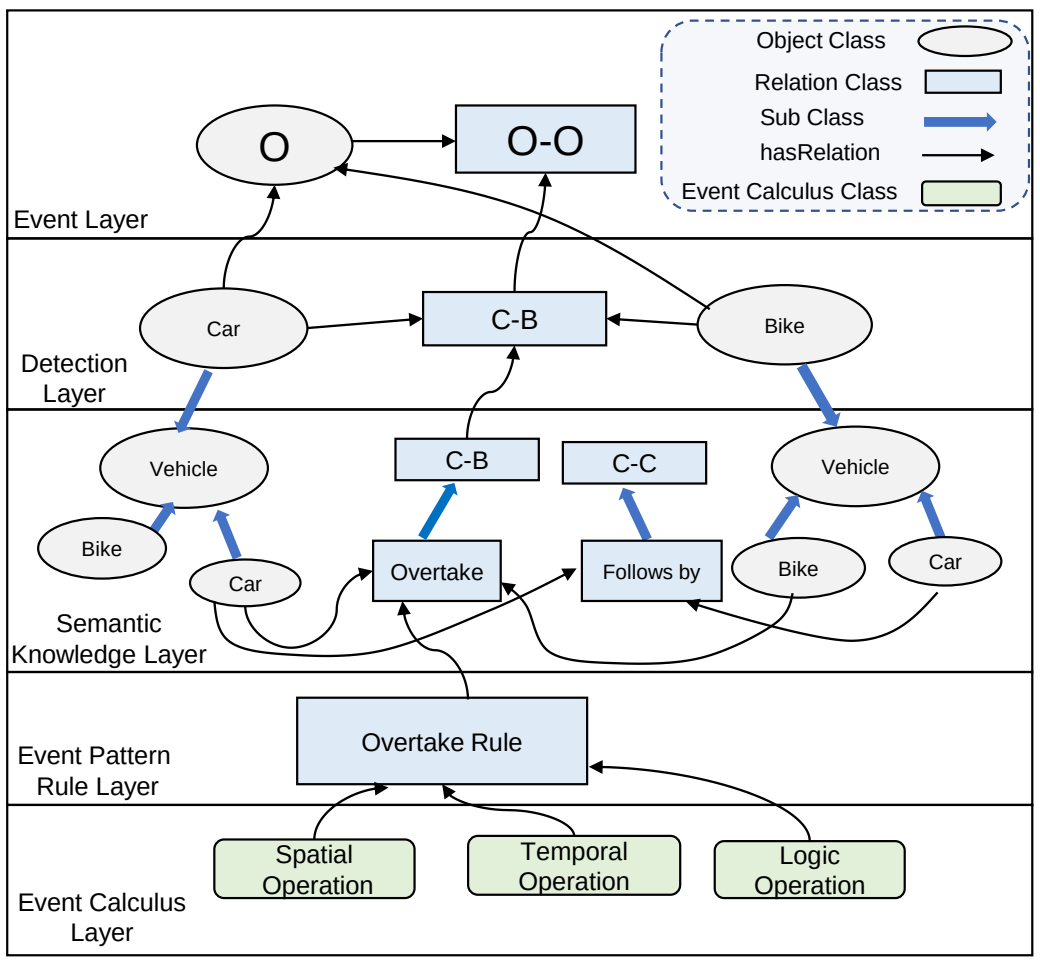

**Fig. 3.** Multimedia Event Relation Network (MERN)

*Event Layer:* This layer consists of Object and O-O Relation which we have defined as an MM event in multimedia CEP engine. The user can create complex event patterns by using MM events and temporal and logical operators.

*Detection Layer:* The Detection layer formulates the multimedia event patterns which the CEP engine can detect. It works in two steps- 1) Objects which the DNN model can detect. For example, as shown in Fig. 3 an object detection model is trained on 'Car' and 'Bike' class. Thus it will detect these two object classes from the streams. 2) To enhance the capability of the engine an O-O relation class is embedded in the layer. This O-O relation class is build using the background knowledge which is retrieved from the Semantic Knowledge Layer. This enhances the capability not only to detect objects using the DNN models, but it can also detect relations among those objects. For example, in Fig. 3 detection layer can detect C-B (car to bike) relation without any additional training of DNN model.

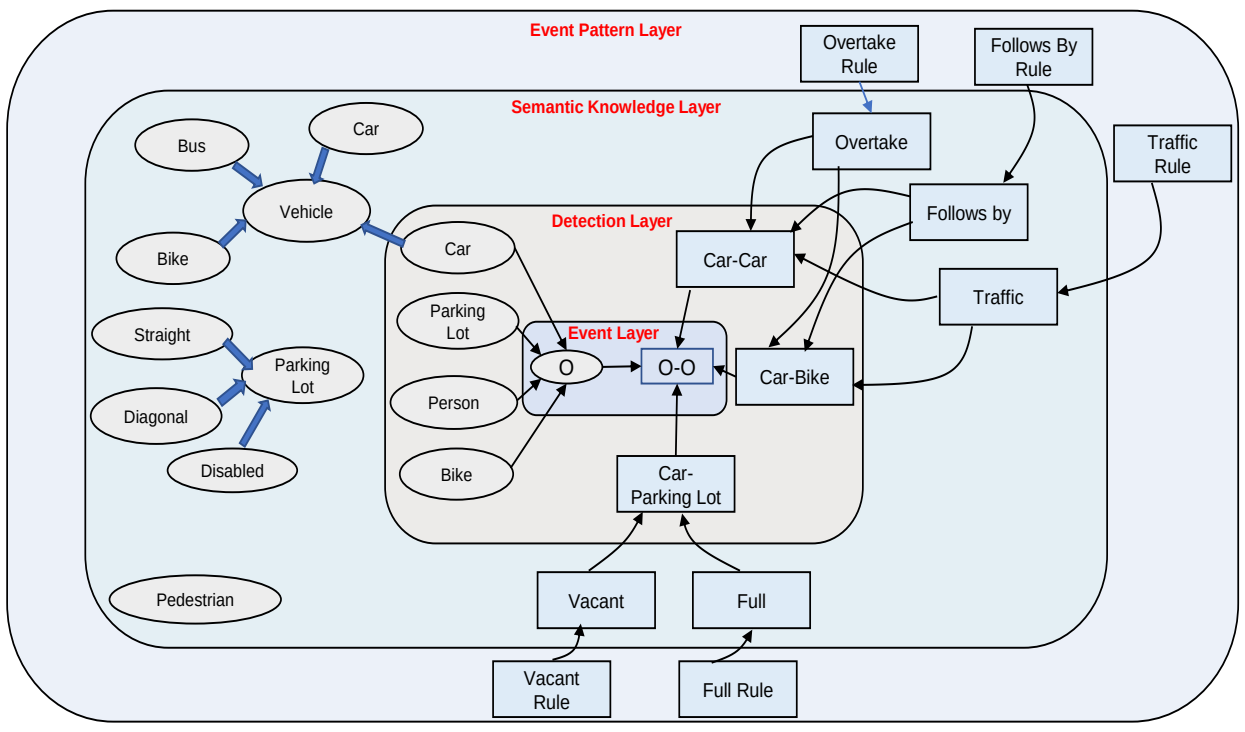

**Fig. 4.** Prototype of Traffic Event MERN Model

*Semantic Knowledge Layer:* This is the background knowledge which is built to enable pattern detection. We consider it as a domain model which detects a relevant pattern within a domain. Fig. 4 shows an example of small traffic events MERN model where patterns like 'overtake', 'follows by' are built. The semantic knowledge layer is built using DNN trained annotations data and further enriched with related background data. To the best of our knowledge, this is a novel concept where an ontology is created with the help of the training annotations of any DNN model. Our semantic knowledge model fetches initial Object (O) classes from the training annotation of DNN model and built O-O relation by domain model expert to enrich it with background knowledge. Thus, the semantic knowledge layer act as a basic backbone of MERN hierarchy. It is a pluggable architecture where new domain models can be created with the help of domain expert and put into the CEP engine enabling a wider range of pattern detection.

*Event Pattern Layer:* The Event pattern layer defines the relationship between various object entities. It uses operators from the event calculus layer to encode different types of relationships and abstract them to human understandable semantic labels. For example, the O-O relation 'Overtake' can be semantically understood by the user and they can create a query over this O-O relation but what 'Overtake' is meant at the machine and the logical level is encoded in this layer.

*Event Calculus Layer:* The event calculus layer encodes all the logic of basic spatio-temporal event calculus which is discussed in section 5.

### 4.3 Architecture

Fig. 5 shows the proposed architecture of multimedia CEP engine with the DNN models embedded in the pre-processor. The engine can be divided into five main components:

*Query and Query Register:* The subscriber can write a query with the help of the MERN model to get information about O-O relations and Objects present in its network. The user can choose that relation pattern and then can built complex event patterns by specifying different publishers over which they want to detect the patterns. The Query Register stores all the queries from a different subscriber which CEP engine needs to detect.

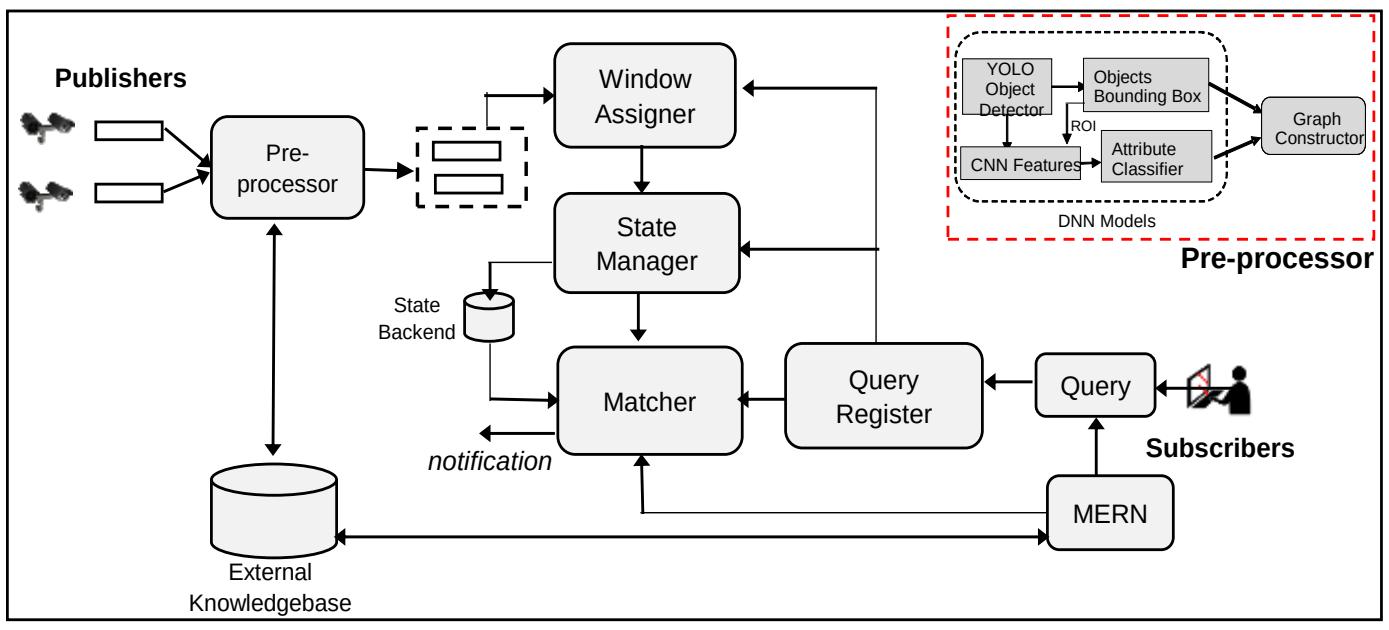

**Fig. 5.** Reference Architecture for Multimedia CEP Engine

*Pre-Processor:* As shown in Fig. 5 (red box), the Pre-processor constitutes of the DNN object and attribute detection model's which are pre-trained on specific datasets. It fetches the data from the publisher data queue and extracts semantic concept like media objects using $V_{COM}$ representation from data streams.

*Window Assigner and State Manager:* The window assigner assigns different window types to different publishers as per the query configuration. Window captures the temporal snapshot of the current stream which is known as the state. The State Manager handles the created window state and sends the state information to state backend and matcher when the window condition is fulfilled.

*Matcher:* Matcher performs spatial and temporal operations over the received state from the state manager. It uses the MERN for detecting a different type of O and O-O relation and then maps them to the registered query to detect the patterns.

# 5 Event Rules and Query Formulation

## 5.1 Event Calculus

The Event calculus is a logic-based programming formalism which allows events to be explicitly represented and is used to build high-level human understandable events [16]. We have used the event calculus to formalise and interlink semantic concepts across multimedia data streams using spatial, temporal and logical relations.

**Spatial Relation:**

Using spatial calculus, we have categorised spatial relations into three main classes:

- *Geometric Representation for Spatial Object ($S_g$):* Fig. 6 shows a spatial object can be represented using geometry-based features like point, line and polygon. We have used bounding boxes-based polygon to represent our objects.
- *Topology-Based Spatial Relation (ST):* We have used Dimensionally Extended nine- Intersection Model (DE-9im), a 2-dimensional topological model to describe pairwise relationships between spatial geometries ($S_g$). The nine relationships its captures are- {*Disjoint, Touch, Contains, Intersect, Within, Covered by, Crosses, Overlap, Inside*} of which four are shown in Fig. 6.

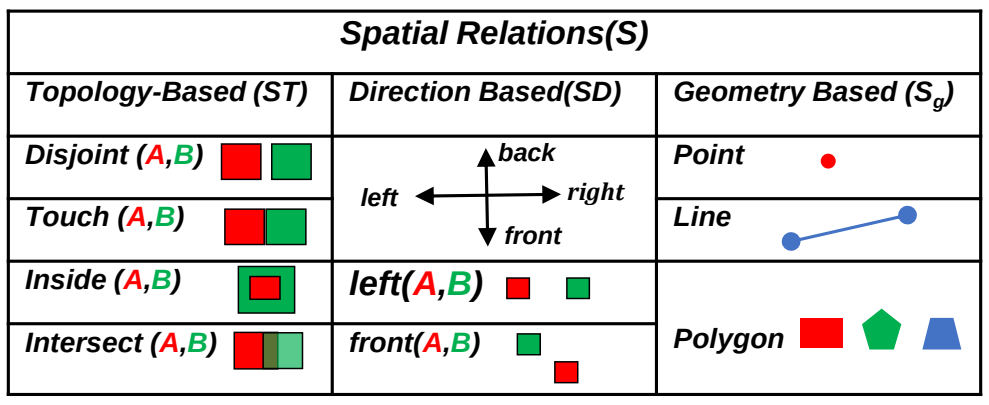

**Fig. 6.** Spatial Relations

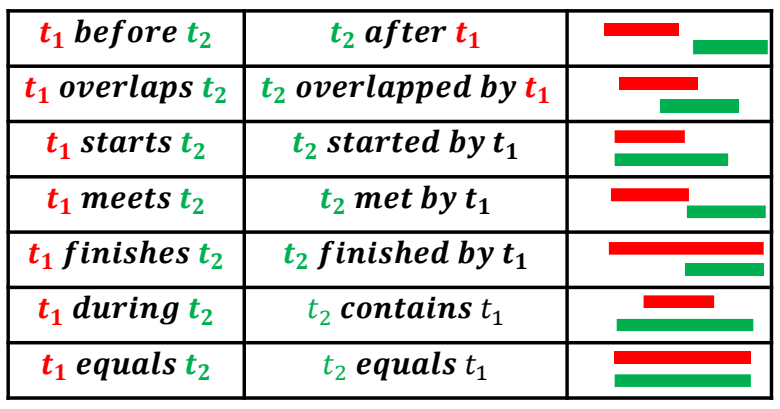

**Fig. 7.** Allen Temporal Relation

- *Direction Based Spatial Relation (SD):* Direction captures the projection and orientation of an object in space. We have used a simpler version of Fixed Orientation Reference System (FORS) [17] which divides the space into four regions: {front, back, left, right}.

To calculate the numerical association between spatial relation we have devised two types of spatial functions-

- *Boolean Spatial Function(bsf):* It returns the boolean value between spatial relation i.e. 0 if relation is false else 1 if relation is true.

$$\boldsymbol{bsf} \rightarrow \boldsymbol{S}(\boldsymbol{Sg_1}, \boldsymbol{Sg_2}) = \begin{cases} 0 \ \ if\ spatial\ relation\ is\ \ false \\ 1\ if\ \ spatial\ relation\ is\ \ true \end{cases}$$

For example, for a topological spatial relation (*ST*) '*Touch*' the boolean spatial function for two objects O1⇝$S_{g1}$ ,and O2⇝$S_{g2}$ will be *bsf (Overlap (O1, O2))* = 0 or 1.

- Metric Spatial Function (msf): msf gives the metric value between the spatial entity- $\boldsymbol{msf} \rightarrow \boldsymbol{\mathcal{M}}(\boldsymbol{g_1}, \boldsymbol{g_2})\ |\boldsymbol{C}|\ \boldsymbol{r}$ where $C \in comparison\ operator$ and $r \in \mathbb{R}$ (real number) and $\mathcal{M} \in any\ \ metric\ calculaltion$ like *msf(DISTANCE(O1,O2)) = 5*.

**Temporal and Logical Relation**

For temporal modelling, we have used the Allen time-intervals [18] (see Fig. 7). Except for the spatial and temporal relation, we have used the logic operators {AND (∧), OR (V), NOT (¬), ANY (∃), EVERY (∀), NOR (↓) , XOR (⨁), XNOR (ʘ), Implies (⟶), Bi-Implies (⟷)}, mathematical and comparison operators {+,-,*, /, < >=} to model the relationships .

## 5.2 Event Rule for 'Overtake'

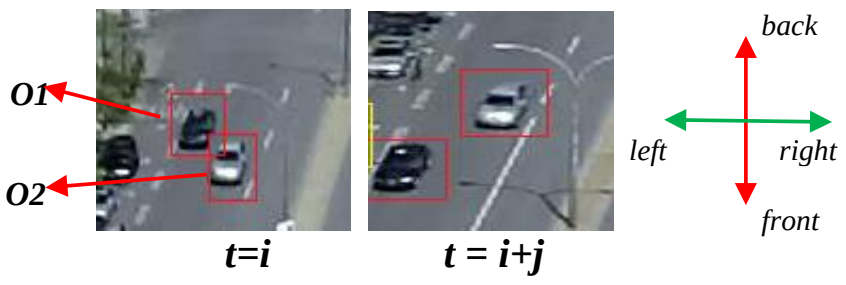

**Fig. 8.** Overtake Scenario

We define 'overtake' as '*change in the relative position of the object in the same direction of motion*'. In Fig. 8 two frames of a video are shown at time frame $t_i$ and $t_{i+j}$ where $t_i < t_{i+j.}$ We see that relative position of the object $o_1$ was 'back' of $o_2$ at $t_i$ which become 'front' at time $t_{i+j.}$ which proves that object $o_1$ has crossed the $o_2$ in i+j$^{th}$ time instance. Thus, we can write this as $\boldsymbol{back}((\boldsymbol{o_1}, \boldsymbol{o_2})^{t_i}) \rightarrow \boldsymbol{front}((\boldsymbol{o_1}, \boldsymbol{o_2})^{t_{i+1}})$. Now, let's deduce the detailed version which will be applied over media stream to detect 'overtake' pattern.

$$\exists(t_i)T\ if\ [bsf(SD(o_1, o_2)x^{t_i}) \odot bsf(SD(o_1, o_2)x^{t_{i+1}})]^{\boxplus[t_m, t_n]}\ where\ x \in back - front\ direction\ \ and\ tm \le ti\ and\ ti + 1 \le tn$$

$$= \begin{cases} 0\ if\ overtake \\ 1\ if\ no\ overtake \end{cases} \qquad \textbf{\textit{(1)}}$$

In equation 1, $\boldsymbol{bsf}(\boldsymbol{SD}(\boldsymbol{o_1}, \boldsymbol{o_2})\boldsymbol{x}^{\boldsymbol{t_i}})$ means a boolean spatial function over spatial direction ($\boldsymbol{SD}$) on objects ($\boldsymbol{o_1}, \boldsymbol{o_2}$) at time instance *i* where we are looking in back -front direction. So, it will evaluate as 'Is $\boldsymbol{o_1}$ back of $\boldsymbol{o_2}$' in back-front direction' which is true, so it will return 1. Similarly, $\boldsymbol{bsf}(\boldsymbol{SD}(\boldsymbol{o_1}, \boldsymbol{o_2})\boldsymbol{x}^{\boldsymbol{t_{i+1}}}$ is the calculation at next time instance i+1 (i.e. next frame). In this case the relative position of object $\boldsymbol{o_1}$ became front of object $\boldsymbol{o_2}$, so '$\boldsymbol{o_1}$ back of $\boldsymbol{o_2}$' become false and thus returns 0. If we do XNOR ($\odot$ )of these two values i.e. 1 at $t_i$ and 0 at $t_{i+1}$ then we get 0 which as per equation (1) means 'overtake'. If there was no change in the relative position of the object in the referred direction, then xnoring will always return 1 which means no overtake occurred between the two objects. We calculate the center of bounding boxes of two objects and then measure the relative back and front of two objects. The evaluation of each frame was done over a window ($\boxplus[t_m, t_n]$) operator. So, for any time instance between this time range if we get 0 between two consecutive frames of objects then we say there is an overtake between these two objects.

### 5.3 Event Rule for Parking Slot Full or Vacant

We define a Parking slot full as '*if the overlap of a queried object over a defined parking slot is greater than some threshold*' then we can say that the object is occupying the parking slot. The parking lot full pattern equation can be written as:

$$\exists\eta slot \in G\ \text{and}\ \forall(O)\ at\ ti \in T\ if\ msf\big(ST(\eta slot, O)\big) > r\ \ where\ ST = overlap\ and\ O = object\ and\ r \in \mathbb{R} \qquad \textbf{(2)}$$

As per equation 2 for any time instance $ti$ if any slot $\eta$ defined over space $G$ overlap with object $O$ with a *msf* value greater than *r* then we can say that the slot is been occupied by the queried object.

## 6 Experimental Setup and Results

### 6.1 Implementation and Datasets

We have implemented the system in Java 8 and performed our experiments on an Intel Core i7 machine with 2.60 GHz CPU and 8 GB of RAM. For initial video preprocessing we have used Java OpenCV library, and for multimedia content retrieval we have used Deeplearning4j which is a java based deep learning library. This is a multithreaded system which can handle multiple video streams in parallel and can detect patterns in them. For object detection, we have used DNN based YOLO [3] model which was pre-trained on PASCAL VOC dataset. For attribute extraction, we fetched the features based on bounding box coordinates from YOLO model layer and then pass it to another CNN classification model trained on specific attributes like color. Table 2 shows the list of some video collected from different datasets. We have created small and cropped video clips of interested event patterns ranging from 5-10 seconds and created a ground truth data manually.

**Table 2.** Dataset Specification

| Video Publisher | Dataset | FPS | Usage |
|---|---|---|---|
| P1 | Pexels | 30.8 | Object Detection |
| P2 | Pexels | 30.25 | Overtake Pattern |
| P3 | VIRAT | 30 | Parking Lot |
| P4 | VIRAT | 32.6 | Parking Lot |
| P5 | ViDVRD | 26.6 | Overtake |
| P6 | Pexels | 30.16 | Overtake |

**Table 3.** Different Subscribers Query

| Subscribers | Query |
|---|---|
| S1 | **Q1** = {[(Car), Ø] $^{\boxplus count\ (5)}$ } |
| S2 | **Q2**= {[(Car $_{Color:\ Black,}$), Ø] $^{\boxplus count\ (5)}$ } |
| S3 | **Q3**= {[(Vehicle), Ø]$^{\boxplus count(5)}$} |
| S4 | **Q4**= {[Ø, ParkingLotFull (Car, Slot)] $^{\boxplus count\ (5)}$} |
| S5 | **Q5**= {[Ø, Overtake (Car, Bike)] $^{\boxplus count\ (5)}$} |

## 6.2 Evaluation Results

Table 3 shows the list of query patterns with different subscribers. The query pattern is listed as per their increasing complexity. In Q1 and Q3 subscriber is only interested in objects while in Q2 the subscriber is interested in both object and its attributes. In Q4 and Q5 the subscriber has queried for a complex spatiotemporal pattern which the system will detect with the help of pattern rules encoded in the MERN model. We performed different experiments on these queries using different publishers (video streams) to understand the efficiency of the engine and proposed MERN model.

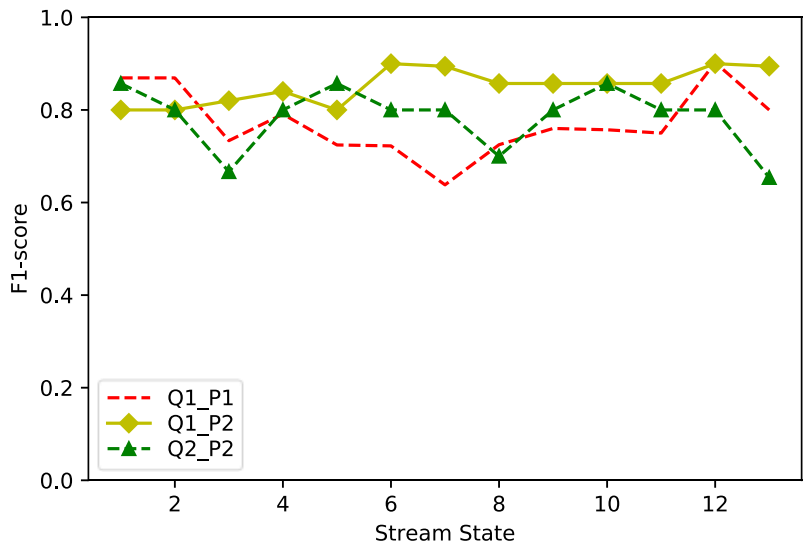

**Fig. 9.** Query Accuracy

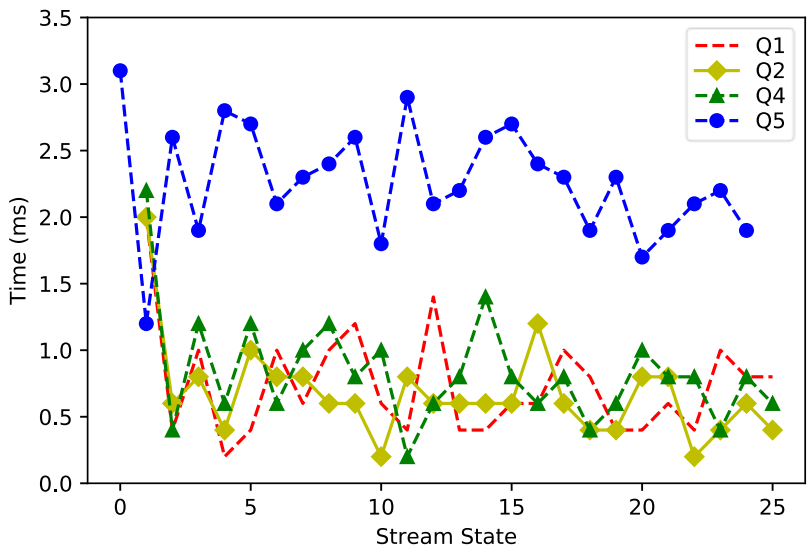

**Fig. 10.** Query Matching Latency

**Quantitative Analysis:**

*Default Scenario:* We have considered Q1 as our default scenario where a subscriber wants to detect only an object (Car). This can be done with or without our engine simply using a DNN based object detector, but our CEP engine enables multiple users to detect objects from various streams. In Q2 since the user is also interested in the attribute of Object (here colour 'Black'), the MERN model activates the attribute CNN model in the pre-processor pipeline and pass the detected object features to get colour value. Fig. 9 shows the F1 score of Q1 and Q2, with the stream state which is captured using windows operator (⊞) and the average F1-score, is calculated across that state. For example, in Q1 the window captures five frames in one state and then average the F-score of each frame across that state. Overall the F1- score of Q1 on stream P1(Q1_P1) is low ( 0.63-0.90) as compare to stream P2(Q1_P2) (0.80-0.90)because in P1 there are multiple fast-moving objects('Car') thus due to a more objects and occlusion there are false positives reducing the overall F1 score of the Q1_P1. Similarly, the F1-score for Q2_P2 ranges between 0.685 to 0.85, which says that it can handle the Object and its attributes classification with good accuracy. We have not calculated the F score for other queries because pattern like 'overtake' are quite rare in datasets and with fewer samples, the accuracy will be biased. For these patterns, we have a qualitative analysis later in this section.
*Matching Latency of Query Pattern:* Fig. 10 shows the average processing time of each state for different query pattern. This is the time when the matcher receives the state and perform detection analysis using MERN model. We have not considered here the pre-processing time of DNN model to detect the Objects and its conversion to $ME_{KG}$ graph. Q1, Q2 and Q3 query act on each image frame, so their response time is good ranging from 0.5 milliseconds to 1.4 milliseconds. Since Q5 tries to detect pattern across image frames, as for 'Overtake' it needs to compare the relative position of the car across frames thus its response time ranges from approximately 1.3 to 3.2 milliseconds. We can see initial spikes at the start due to the extra time added in the state formation as DNN model load into the memory. In experiments stream state on x-axis means that the experiment was performed on 125 (25*5) image frames($ME_{KG}$ graph).
*System Throughput:* Throughput means the number of images the system can process per second. As shown in Fig. 11 the system can process maximum 7.6 frames/sec when it consumes three video streams parallelly on CPU performance. After this, the system throughput starts decreasing due to memory overhead as several publishers start loading the computationally intensive DNN models in the memory.

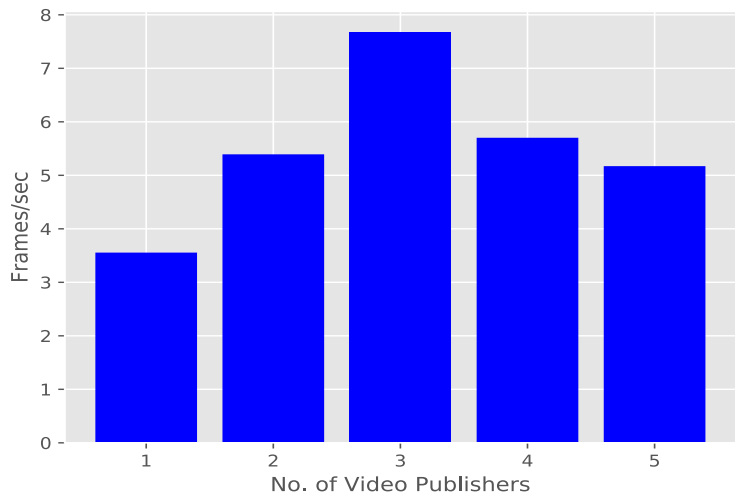

**Fig. 11.** System Throughput

**Table 4.** Comparison of DNN Object Detector vs MERN Enabled CEP

| Query Support | DNN Object Detector Only | DNN+ MERN Enabled CEP |
|---|---|---|
| Q1 | Yes | Yes |
| Q2 | No | Yes |
| Q3 | No | Yes |
| Q4 | No | Yes |
| Q5 | No | Yes |

**Qualitative Analysis:**

Table 4 shows the comparison of queries supported by DNN object detectors and MERN enable CEP. If we see that Q3 ('Vehicle') is also an object detection query but the DNN model is unable to detect it, as it was not trained on the Vehicle training annotation. This limitation can be easily handled by semantic enrichment using background knowledge. Since the MERN has already background relationship that the DNN model can detect a 'Car' from its detection layer capability and 'Car' is a subclass of the 'Vehicle' which was already embedded in its semantic knowledge layer. Thus, it was able to detect Q3 using simple background enrichment.

Fig. 12 shows the frames of publisher P2 at different time instances. The red dots show the trajectory of each object at different time instances. It can be easily visualised from the third frame that left car overtakes the right one. This was calculated using the overtake pattern rule which was built in MERN model. Similarly, Fig. 13 shows the intersection of the car with the parking slot in different time instances. In the last frame the overlap between car and parking slot was maximum, and thus the subscriber (S4) got the notification. Here the calculation is done on the basis of state. The matcher looks the state and sends notification if the car would have moved from the slot then the matcher will send parking lot empty information to the subscribed user in near real time. Fig. 14 shows a 'overtake' pattern miss although both objects were in the same direction. This is due to moving camera and its angle of projection to take the video.

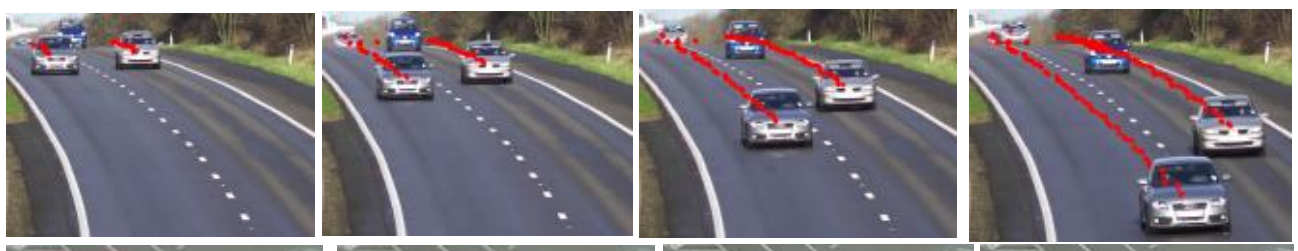

**Fig. 12**'Overtake' Pattern Detection on Road

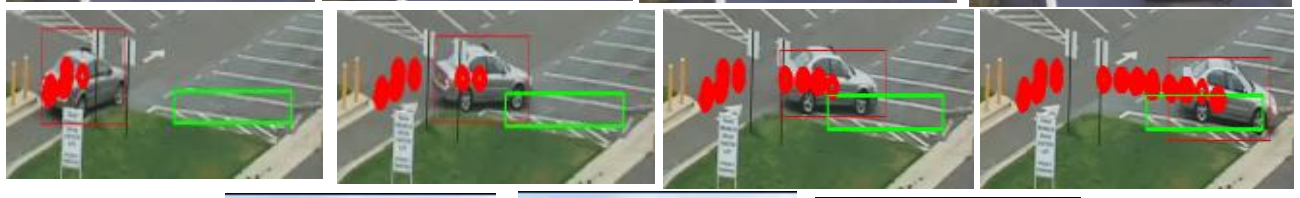

**Fig. 13.** Parking Lot Full Event Detection

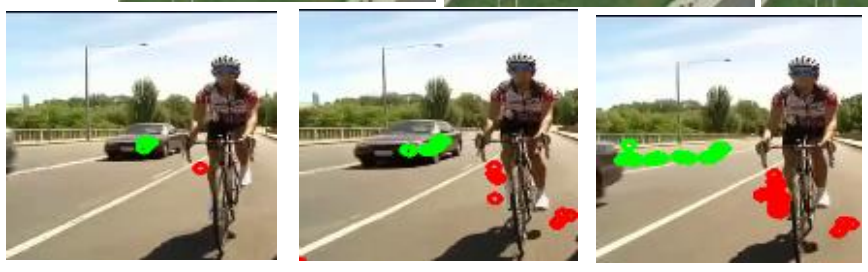

**Fig. 14.** 'Overtake' pattern miss due to the moving camera.

### 6.3 Limitations

Our work has certain limitations and assumptions which are as follows- 1) Multimedia objects are highly complex and can have multiple relationships with each other. Our initial work has considered only O-O relation which is 'one to one' mapping between two objects instead of 'one to many' or 'many to many' relations. 2) We have considered all our calculations in 2-dimension plane while in the real world the relations are quite complex and spread in 3-dimensions, leading to many patterns misses or false event detection. An example can be seen in Fig. 14 where it was an 'overtake' pattern miss. 3) DNN models are basic building blocks for our CEP, any prediction failure in them will decrease our CEP engine performance. 4) DNN models are very brittle and dependent on training data. Our MERN model enhances the performance of these DNN models using an ensemble of patterns. The training data on which the model is trained limits what CEP system can detect.

## 7 Conclusion and Future Work

The main aim of our work is to detect patterns in CEP from low-level features in the multimedia stream using semantic representation. We have proposed a semantic representation of multimedia streams using the $V_{COM}$ ontology which helps to structure low-level media data for pattern reasoning. We have proposed a Multimedia Event Relation Network (MERN) which can help in creating multiple ensembles of patterns by aligning DNN model prediction capability. The paper details the reference architecture for the multimedia CEP engine and calculated the F1-score, pattern latency and system throughput to analyse the performance of the CEP engine. In future, we will focus on how to handle representations for One to Many and Many to Many O-O relations and optimise the overall system performance.

## Acknowledgement

This work was supported with the financial support of the Science Foundation Ireland grant 13/RC/2094 and co-funded under the European Regional Development Fund.